\documentclass[letterpaper, 10 pt, conference]{ieeeconf} 
\usepackage{cite}
\usepackage{amsmath}

\usepackage{algorithmic}
\usepackage{algorithm}

\usepackage{array}
\usepackage{url}

\usepackage{amssymb}  
\usepackage{booktabs}
\usepackage{multirow}
\usepackage{subfig} 
\usepackage{newfloat}

\usepackage{times}  
\usepackage{helvet}  
\usepackage{courier}  
\usepackage{caption} 
\usepackage{algorithm}
\usepackage{algorithmic}
\usepackage{amssymb}
\usepackage{amsmath}
\usepackage{multirow}
\usepackage{booktabs}
\usepackage{newfloat}
\usepackage{listings}
\usepackage{graphicx}

\begin{document}

\title{DenseLiDAR: A Real-time Pseudo Dense Depth Guided Depth Completion Network}

\author{Jiaqi Gu$^{1}$, Zhiyu Xiang$^{2}$, Yuwen Ye$^{1}$ and Lingxuan Wang$^{1}$
%
\thanks{The work is supported by NSFC-Zhejiang Joint Fund for the Integration of Industrialization and Informatization under grant No. U1709214 and Key Research \& Development Plan of Zhejiang Province (2021C01196). } 
\thanks{$^{1}$Jiaqi Gu, Yuwen Ye and Lingxuan Wang are with College of Information Science \& Eletronic Engineering, Zhejiang University, Hangzhou, China, {\tt\footnotesize\{vadin, yuwenye, 21960188\}@zju.edu.cn}}%
\thanks{$^{2}$Zhiyu Xiang, corresponding auther, is with Zhejiang Provincial Key Laboratory of Information Processing, Communication and Networking, Zhejiang University, Hangzhou, China.{\tt\footnotesize xiangzy@zju.edu.cn}}%
}
\maketitle

\begin{abstract}

Depth Completion can produce a dense depth map from a sparse input and provide a more complete 3D description of the environment. Despite great progress made in depth completion, the sparsity of the input and low density of the ground truth still make this problem challenging. In this work, we propose DenseLiDAR, a novel real-time pseudo-depth guided depth completion neural network. We exploit dense pseudo-depth map obtained from simple morphological operations to guide the network in three aspects: (1) Constructing a residual structure for the output; (2) Rectifying the sparse input data; (3) Providing dense structural loss for training the network. Thanks to these novel designs, higher performance of the output could be achieved. In addition, two new metrics for better evaluating the quality of the predicted depth map are also presented. Extensive experiments on KITTI depth completion benchmark suggest that our model is able to achieve the state-of-the-art performance at the highest frame rate of 50Hz. The predicted dense depth is further evaluated by several downstream robotic perception or positioning tasks.  For the task of 3D object detection, 3\~{}5 percent performance gains on small objects categories are achieved on KITTI 3D object detection dataset. For RGB-D SLAM, higher accuracy on vehicle’s trajectory is also obtained in KITTI Odometry dataset. These promising results not only verify the high quality of our depth prediction, but also demonstrate the potential of improving the related downstream tasks by using depth completion results. 

\end{abstract}

\begin{figure}[!t]
  \centering
  \subfloat[RGB Image]{
    \centering
    \includegraphics[width=0.22\textwidth]{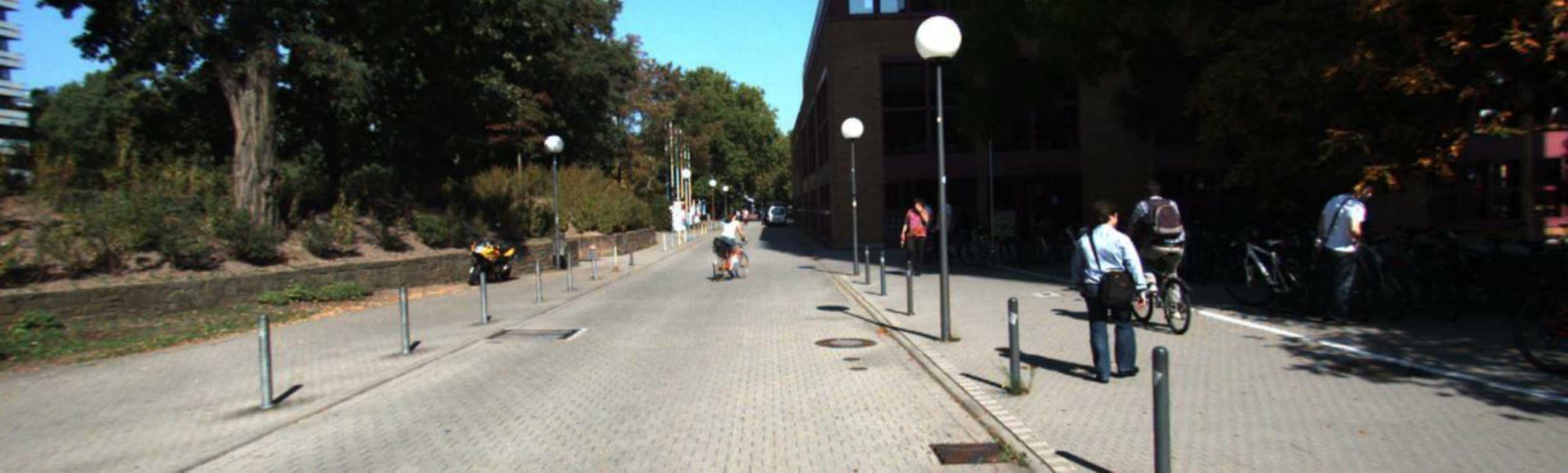} }
  \subfloat[Sparse Depth]{
    \centering
    \includegraphics[width=0.22\textwidth]{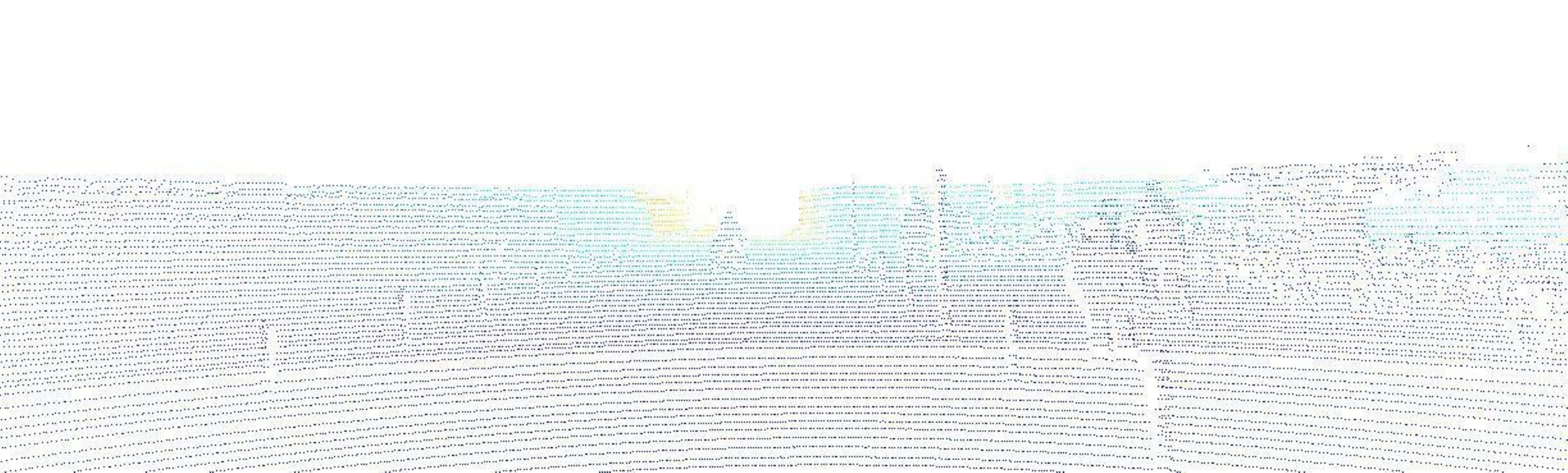} }

  \subfloat[DeepLiDAR\cite{qiu2019deeplidar} Prediction]{
    \centering
    \label{fig1:deeplidar1}
    \includegraphics[width=0.22\textwidth]{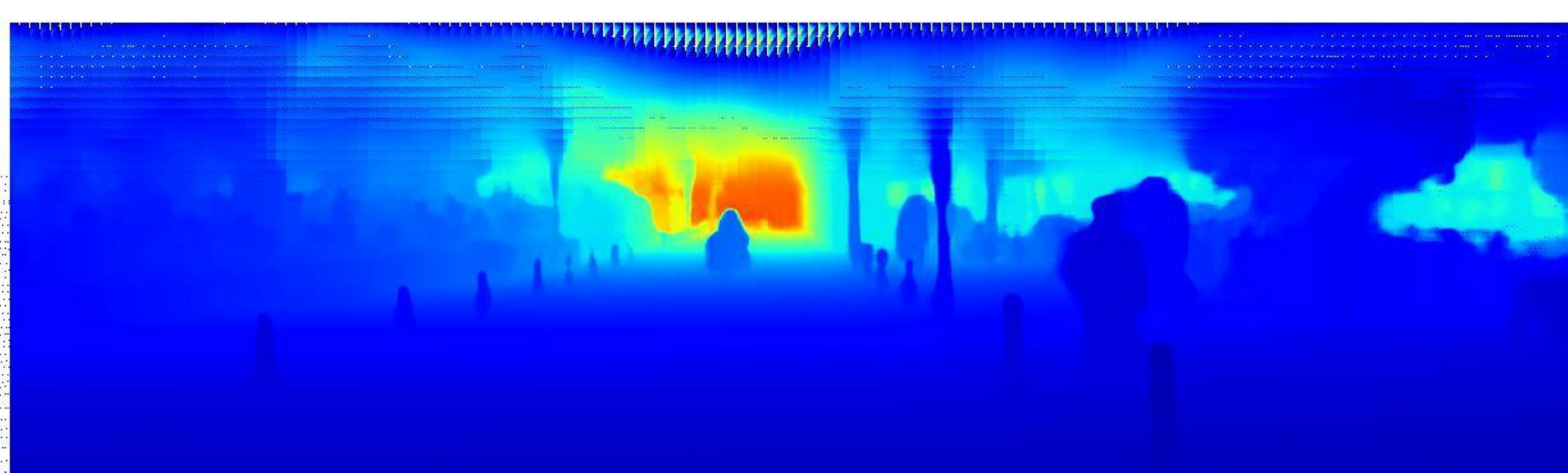}}
  \subfloat[Our Depth Prediction]{
    \centering
    \label{fig1:denselidar1}
    \includegraphics[width=0.22\textwidth]{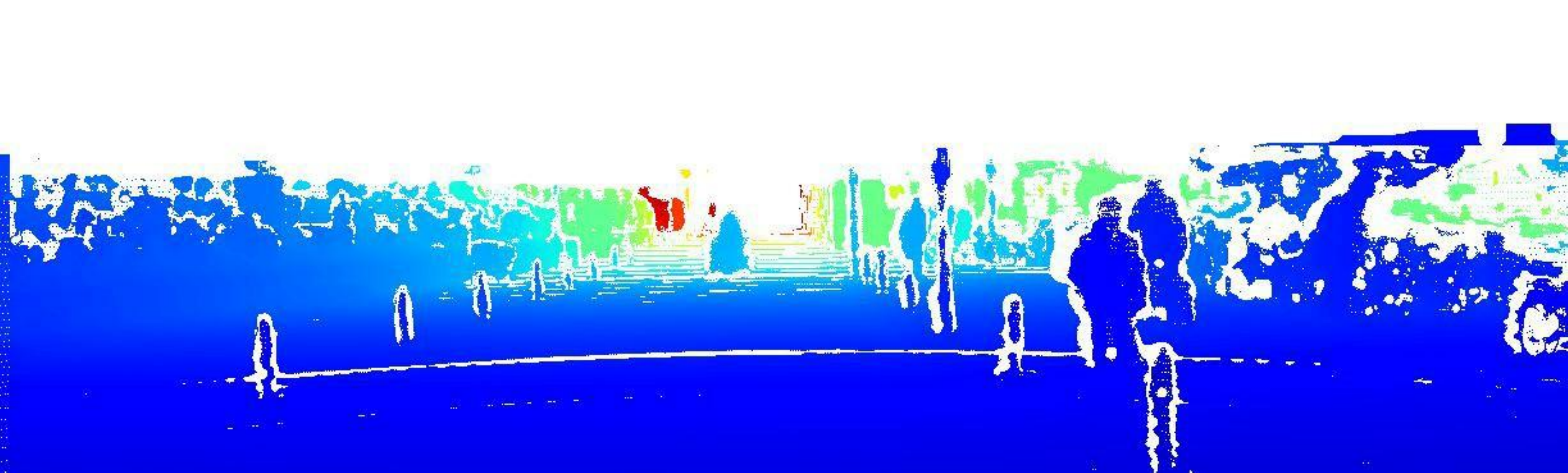}}

  \subfloat[DeepLiDAR\cite{qiu2019deeplidar} Point Cloud]{
    \centering
    \label{fig1:deeplidar2}
    \includegraphics[width=0.22\textwidth]{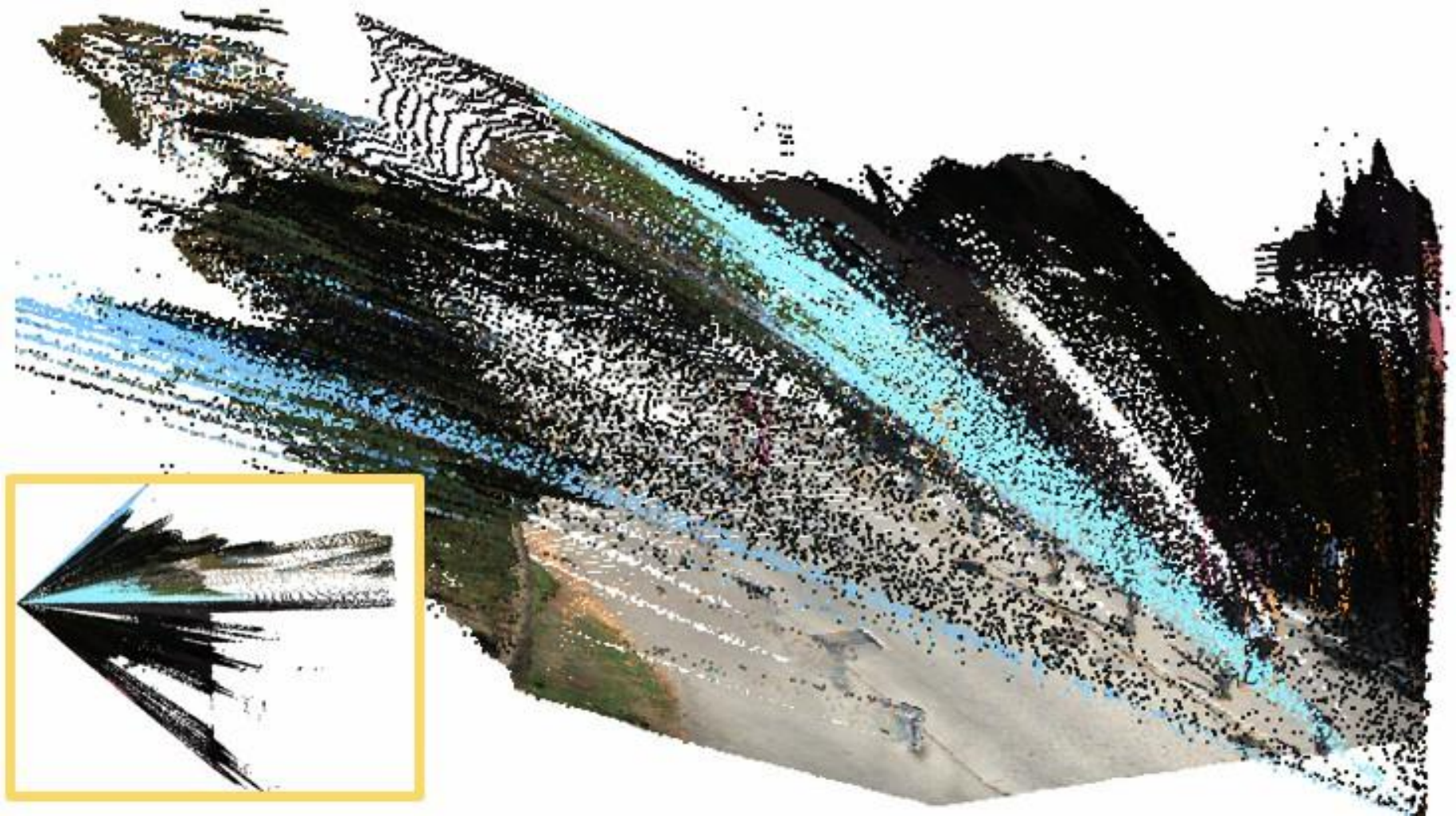}}
  \subfloat[Our Dense Point Cloud]{
    \centering
    \label{fig1:denselidar2}
    \includegraphics[width=0.22\textwidth]{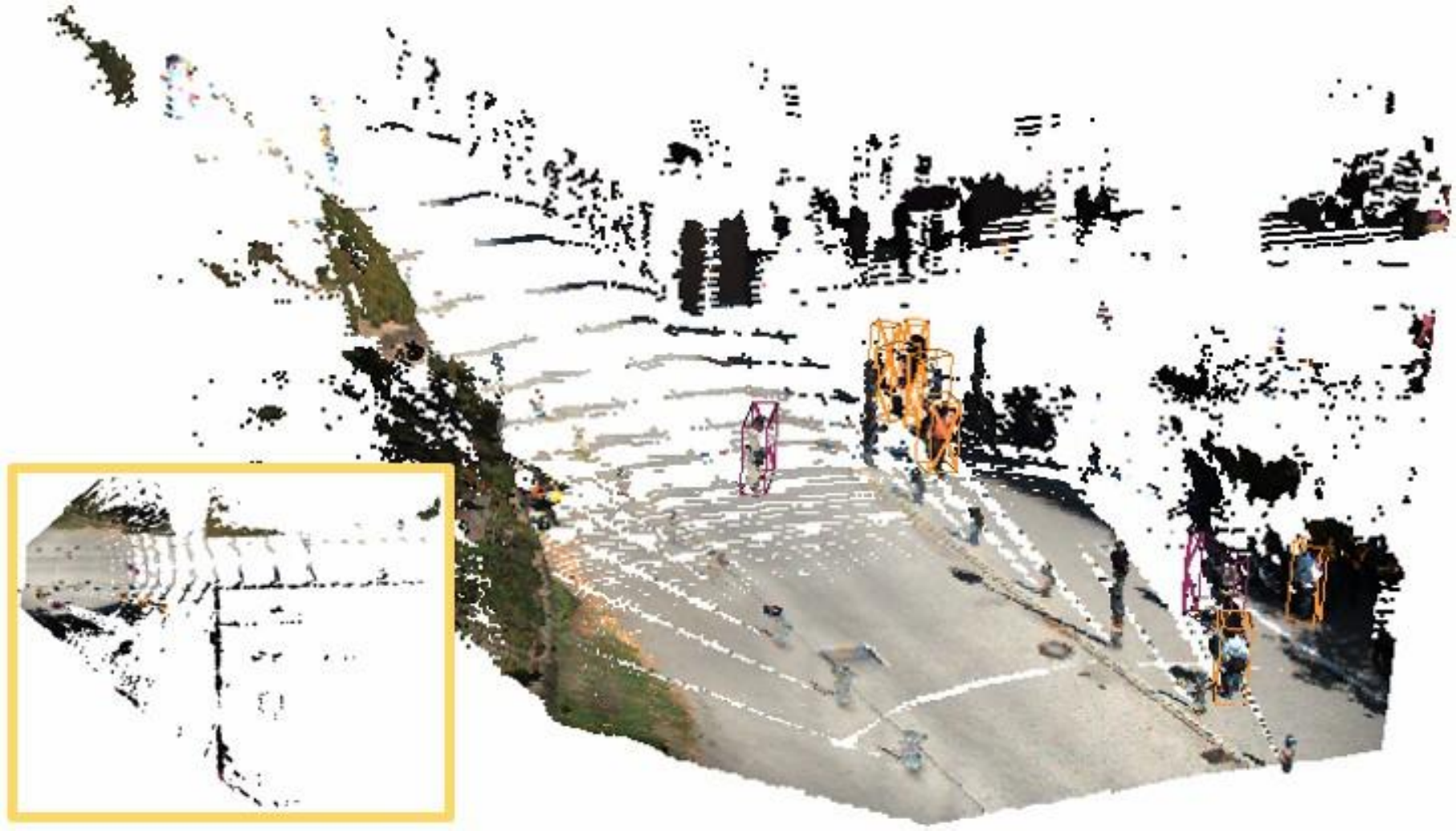}}
  \caption{
  \textbf{ Comparison of Depth Completion Results.} Both DeepLiDAR\cite{qiu2019deeplidar} and our method use RGB image (a) and sparse LiDAR depth (b) to predict dense depth map. Comparing with DeepLiDAR\cite{qiu2019deeplidar} (c,e), our depth map (d) and corresponding point cloud (f) have less erroneous points and more clear object boundaries. The bird-eye-view images marked with yellow bounding box in the last row also demonstrate the high quality of our prediction.}
  \label{fig1:pointcloud}
\end{figure}

\section{Introduction}

Accurate 3D information is vital for lots of computer vision tasks such as AR/VR and robotic navigation. While today’s LiDAR sensors can acquire reliable 3D measurements from the surrounding environment, the resulting depth image is still very sparse when comparing to a medium resolution RGB image (about 3\~{}4\% density). Depth completion refers to the task of estimating dense depth image from a sparse depth input, which aims at providing a more complete 3D description of the environment. With the rapid development of deep learning, great progress has been made in this field. The state-of-the-art methods could produce a dense depth map at around 750mm in root mean square error (RMSE) in KITTI benchmark\cite{uhrig2017sparsity}.

Despite low RMSE achieved, lots of defects still exist in the predicting results. Especially when we observe the completed depth map in 3D space, the point cloud is usually polluted by many outliers which appear to be long-tail or mixed-depth confusion areas, as shown in Fig \ref{fig1:deeplidar1}(e). This phenomenon partly attributes to the sparsity of the  ground truth. Taking the popular KITTI depth completion dataset as an example, the average density of ground truth is only about 14\%, which means most pixels cannot be sufficiently supervised during training. Similarly, when evaluating the prediction, only the pixels with ground truth are evaluated. This makes the RMSE metric upon such sparse positions tends to be over-optimistic. Meanwhile, it is more difficult to apply effective structure constraint on the output given only sparse ground truth. Insufficient structure regulation usually leads to blur depth predictions on object boundaries.

In this paper, a novel real-time pseudo-depth guided depth completion method is proposed. We believe a dense reference depth map is essential for producing a high-quality dense prediction. We generate a pseudo depth map from sparse input by traditional morphological operations. Although not accurate, the dense pseudo depth contains more structural information of the scene and can be very helpful to this task. Specifically, the pseudo depth map works in our model in three aspects: (1) The entire model can be transformed into a residual structure, predicting only the residual depth upon the pseudo map; (2) Some erroneous points caused by the penetration problem in raw sparse input could be eliminated; (3) A strong structural loss enforcing the sharp object boundaries could be applied. Besides the novel model for depth completion, two new evaluation metrics based on RMSE, \textit{i}.\textit{e}., RMSE\_GT+ and RMSE\_Edge, are also proposed to better representing the true quality of the prediction. The former evaluates the prediction result on a denser ground truth while the latter is mainly focus on the edge quality of the prediction. Experimental results on KITTI depth completion dataset demonstrates the superior performance of our model. An example of our result is shown in Fig \ref{fig1:pointcloud}. To explore the benefits of utilizing the predicted dense depth on downstream robotic tasks, we further apply our results on 3D object detection and RGB-D SLAM. As expected, great improvements are achieved comparing with the sparse input or dense depth from the baseline algorithm. 

In summary, the contributions of this paper are: 

\begin{itemize}
\item We propose a novel pseudo depth guided depth completion neural network. It is designed to predict residual upon pseudo depth, making the prediction more stable and accurate. Data rectification and 3D structure constraint are also guided by pseudo dense depth.
\item We propose two new metrics for better evaluating the quality of the predicted dense depth. RMSE\_GT+ computes the depth error on a carefully complemented ground truth while RMSE\_Edge evaluate the accuracy on edge areas of the depth map. Both metrics are more sensitive than RMSE. Combining these metrics together, more comprehensive evaluation of the quality can be obtained.
\item The entire network is implemented end-to-end and evaluated in KITTI datasets. Experimental results show that our model achieves comparable performance to the state-of-the-art methods at the highest frame rate of 50Hz. Further experiment on using the dense depth prediction on object detection and SLAM tasks also verifies the significant quality improvement of our method and demonstrates the potential of using high quality depth completion in related downstream tasks. 
\end{itemize}

\section{Related Works}
Some early depth completion methods rely on template dictionary to reconstruct the dense depth, such as compressive sensing\cite{hawe2011dense} or wavelet-contourlet dictionary\cite{liu2015depth}. IP-basic\cite{ku2018defense} proposed a series of morphological operations such as dilation and hole filling to densify sparse depth in real time. Despite fast speed, limited performances are obtained for these traditional methods.

Recently, deep learning approaches are leading the mainstream of depth completion. In \cite{6618993}, repetitive structures are used to identify similar patches across different scales to perform depth completion. A sparsity invariant CNN architecture is proposed in \cite{uhrig2017sparsity}, which explicitly considers the location of missing data during the convolution operations.

Inducing RGB image in depth completion task is helpful since RGB images contain abundant scene information which could be an effective guidance for completing the sparse depth \cite{qiu2019deeplidar}\cite{ma2019self}\cite{zhong2019deep}\cite{lee2020deep}. Sparse-to-dense\cite{ma2019self} proposes a self-supervised network which exploits photometric loss between sequential images. CFCNet\cite{zhong2019deep} learns to capture the semantically correlated features between RGB and depth information. CG-Net\cite{lee2020deep} proposes a cross guidance module to fuse the multi-modal feature from RGB and LiDAR. 

Some methods rely on iterative Spatial Propagation Network (SPN) to better treat the difficulties made by the sparsity and irregular distribution of the input [10][11][12][13]. A convolutional spatial propagation network is first developed in CSPN\cite{cheng2018depth} to learn the affinity matrix for depth prediction. CSPN++\cite{CSPNplus} improves this baseline by learning adaptive convolutional kernel sizes and the number of iterations in a gated network. DSPN\cite{xu2020deformable} utilizes deformable convolutions in spatial propagation network to adaptively generate receptive fields. NLSPN\cite{park2020non} expands the propagation into non-local fields and estimates non-local neighbors’ affinities for each pixel. Despite impressive results, these methods have to be trained and propagated iteratively, which are hard to implement end-to-end and run in real time.

Due to the sparsity of the input and the ground truth, ambiguous depth prediction in object boundaries or backgrounds are very common for image-view based depth completion. Some works induce spatial point cloud as an extra view for effective feature extraction \cite{learning2019yun}\cite{hekmatian2019conf}\cite{depth-coefficients-for-depth-completion}. UberATG\cite{learning2019yun} applys 2D and 3D convolutions on depth pixels and 3D points respectively to extract joint features. Conf-Net\cite{hekmatian2019conf} generates high-confidence point cloud by predicting dense depth error map and filtering out low-confidence points. DC-Coef \cite{depth-coefficients-for-depth-completion} transforms continuous depth regression into predicting discrete depth bins and applies cross-entropy loss to improve the quality of the point cloud. PwP\cite{yan2019completion} generates the normal view via principal component analysis (PCA) on a set of neighboring 3D points from sparse ground truth.

\begin{figure*}[h]
  \centering
  \includegraphics[width=16cm]{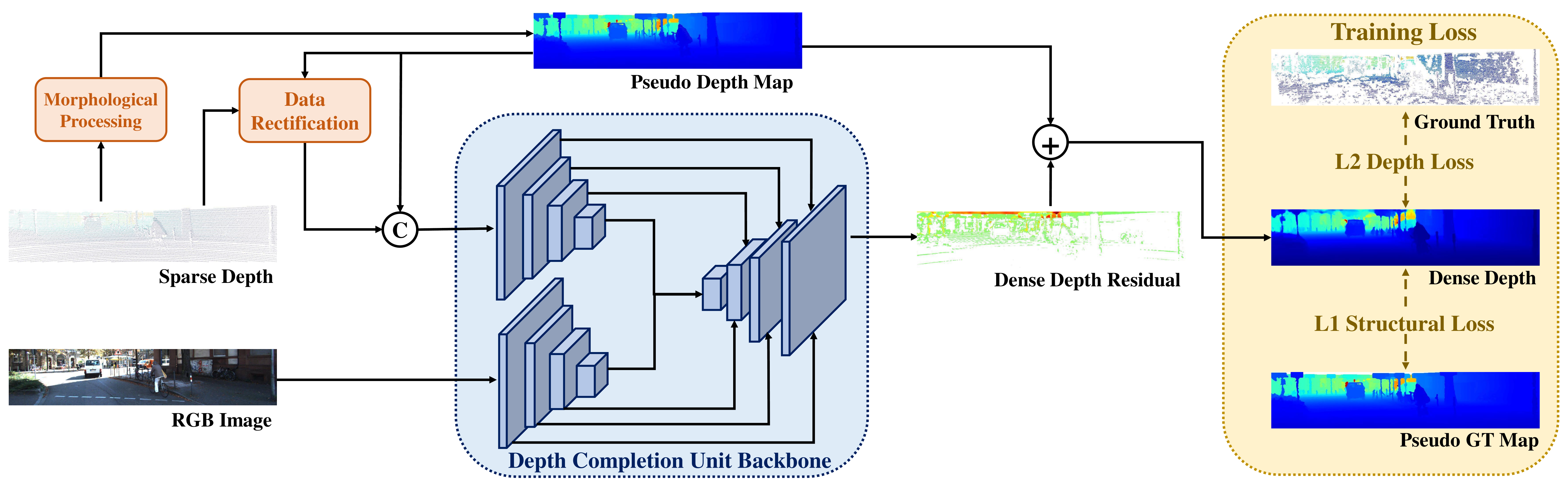} 
  \caption{\textbf{The network structure of our DenseLiDAR model.} Given a pair of sparse depth and RGB image, a dense pseudo depth map is first computed by morphological processing. With the reference of pseudo depth, the raw sparse depth input can further be rectified before feeding into network. The network backbone employs only one DCU\cite{qiu2019deeplidar}, taking the RGB, the concatenation of rectified sparse and pseudo depth images as input and producing a depth residual upon the pseudo depth. During training, both depth loss and structural loss are applied to regulate the prediction.}
  \label{fig2:pipeline}
\end{figure*}

Some methods adopt pre-trained modules from external training data in their network to acheive better performance. RGB-GC\cite{van2019sparse} uses pre-trained semantic segmentation model on Cityscapes dataset\cite{cordts2016cityscapes}. SSDNet\cite{zou2020simultaneous} and RSDCN\cite{zou2020rsdcn} exploit the Virtual KITTI dataset\cite{gaidon2016virtual} to learn the semantic labels of each pixel in dense depth. DeepLiDAR\cite{qiu2019deeplidar} jointly learns the depth completion and normal prediction tasks where the normal prediction module is pre-trained on a synthetic dataset generated from CARLA simulator\cite{2017CARLA}.

In contrast to the methods above, this paper characters employing pseudo depth map which can be easily generated from simple morphological operations to guide and regularize the depth completion task. Our model is built upon a simple network backbone and can be trained by KITTI depth completion dataset without any pre-training. High accuracy and real-time performance could be achieved thanks to the effective guidance of pseudo depth.

\section{Method}

In this section, we first introduce the detailed structure of our DenseLiDAR model, including the residual structure, raw data rectification and loss function of the network. Then two new evaluation metrics are proposed to better evaluate the quality of the prediction. The overall structure of the proposed network is illustrated in Fig \ref{fig2:pipeline}. 


\subsection{Pseudo Depth Based Residual Prediction}

Most depth completion methods often produce ambiguous depth in transition areas between foreground and background. This is partly due to the sparsity of the input and the properties of 2D convolutions used in the network.  It’s hard to regress sharp depth changes (\textit{e}.\textit{g}., regressing a ‘1’ among eight ‘0’s in a 3x3 grid), since 2D convolutions and L2 loss functions tend to regress mid-depth values in edge areas \cite{depth-coefficients-for-depth-completion}. We seek a new way of depth regression that can not only predict accurate depth values, but also produce sharp object boundaries. Our solution is to transform the depth completion task from an absolute value regression to a residual depth regression problem. We first produce a dense pseudo depth map for a given sparse depth input by traditional method, then take it as a reference to construct a residual structure for the network, as shown in Fig \ref{fig2:pipeline}.

\textbf{Pseudo Depth Map.} Pseudo depth map is generated from sparse input by fast morphological steps such as kernel-based dilation, small hole closure and large hole filling \cite{ku2018defense}. The resulting dense pseudo depth map contains sharp edges and rich structural information, which makes it very suitable to be a coarse reference for the final output.

\textbf{Residual-based Regression.} As suggested in Fig \ref{fig2:pipeline}, our network does not directly regress the absolute depth value for each pixel, but predicts the residual depth upon the dense pseudo depth map. Comparing with the former solution, residual-based regression has two advantages:

1) The output is pre-normalized to each pixel. In conventional solution, pixels with larger depth tend to be penalized more due to the usage of L2 loss in training. By predicting the residual depth, the unbalanced loss distribution problem caused by absolute depth is largely mitigated. 

2) Given the coarse pseudo depth, boundary pixels with large residual will be more focused by L2 loss, which is helpful for producing more accurate predictions on object boundaries. In other words, it makes the prediction more structure-adaptive and decreases the mix-depth errors. 

\begin{figure}[!t]
  \centering
  \includegraphics[width=0.45\textwidth]{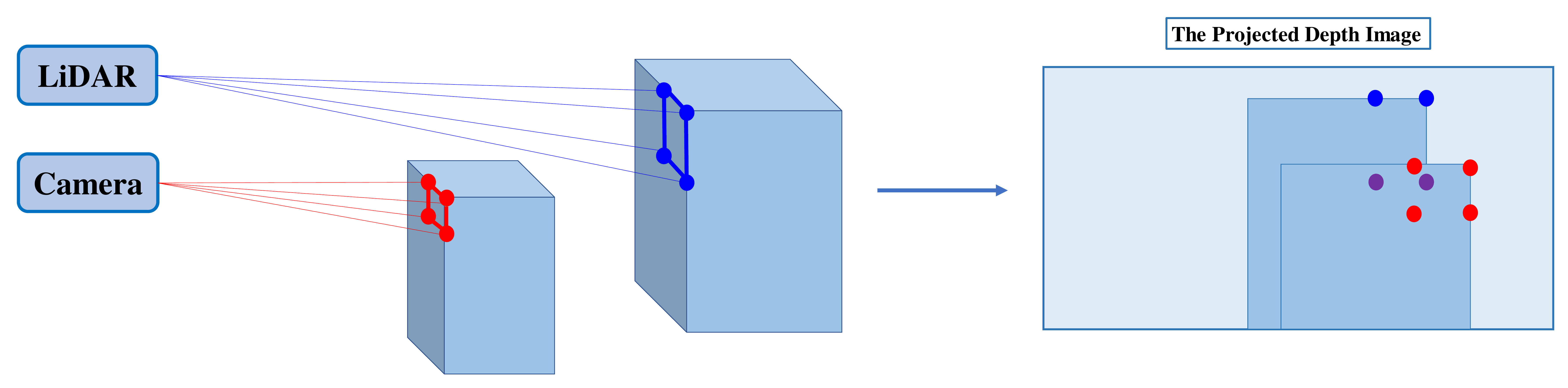} 
  \caption{
  \textbf{Penetration problem in projecting 3D points into depth map.} Because of the displacement between LiDAR and camera, some distant points (marked blue) may penetrate through close objects (marked red) in the projected depth image, resulting in some erroneous depth pixels (marked purple) in the close object.}
  \label{fig3:penetration}
\end{figure}

\subsection{Data Rectification}

\textbf{Penetration Problem} Due to the displacement between RGB camera and LiDAR, projecting sparse LiDAR points to the image view may cause mixed-depth pixels between foreground and background. Those ambiguous areas are mainly around object boundaries, where distant points may penetrate through and appear within the close object on the projected depth image, as illustrated in Fig \ref{fig3:penetration}.

We seek to solve the penetration problem from the source with the help of dense pseudo depth map. Comparing to the raw sparse depth, the pseudo map is relatively immune to the mix-depth problem because small “holes” caused by mix-depth pixels will be filled by nearby foreground pixels during dilation operation. Therefore, we can rectify the raw sparse input by removing those pixels whose difference with the pseudo depth are larger than a threshold. Ablation studies in Experiments part will show the effectiveness of this rectification strategy. 

Concatenated with the pseudo depth map, the rectified sparse depth as well as the RGB image is fed into the network which is composed of a single Depth Completion Unit (DCU) \cite{qiu2019deeplidar} to produce residual dense depth prediction.

\subsection{Loss Functions}
In order to address more attention to shape and structure, besides standard depth loss, we seek to define a structural loss that penalizes depth distortions on scene details, which usually corresponds to the boundaries of objects in the scene. Such kind of loss are popular in leaning tasks when dense supervision are available, \textit{e}.\textit{g}., single image depth estimation\cite{hu2019revisiting}\cite{alhashim2018high}\cite{Ummenhofer_2017}. As ground truth depth is also sparse and lacks of structure and shape details, we rely on pseudo ground truth (GT) map for structure supervising. Pseudo GT map follows the same computing process of pseudo depth map, but is generated from sparse ground truth. 

For training our network, we define the total loss $L$ as the weighted sum of two loss functions, as shown in Eq (1):

\begin{equation}
\setlength{\abovedisplayskip}{-0.2cm}
\begin{aligned}
L(D,\hat{D}) &= L_{depth}(D,\hat{D}+\tilde{D}) + \lambda L_{structural}(\overline{D},\hat{D}+\tilde{D})  \\
\end{aligned}
\vspace{-0.2cm}
\label{eq_loss}
\end{equation} 

where $D$, $\tilde{D}$, $\overline{D}$ and $\hat{D}$ are the ground truth depth, the dense pseudo depth, the dense pseudo GT and the residual depth prediction, respectively. The first loss term $L_{depth}$ is the pixel-wise L2 loss defined on the residual depth:


\begin{equation}
  \setlength{\abovedisplayskip}{-0.2cm}
  \begin{aligned}
  L_{depth}(D,\hat{D}+\tilde{D}) = \frac{1}{n} \sum_{i=1}^n (|D_i -\tilde{D_i} - \hat{D_i}|)^2 
  \end{aligned}
  \vspace{-0.2cm}
  \label{eq_depth}
\end{equation}

The second loss term $L_{structural}$ is composed of two commonly used loss in image reconstruction tasks, i.e., $L_{grad}$ and $L_{SSIM}$, as shown in Eq (3)-(5). The former enforces smoothness of the depth based on the gradient $\nabla$ of the depth image. The latter is the structural similarity constraint SSIM\cite{wang2004image} which consists of luminance, contrast and structure differences to the pseudo GT map.

$$
\setlength{\abovedisplayskip}{3pt}
\setlength{\belowdisplayskip}{3pt}
L_{structural}(\overline{D},\hat{D}+\tilde{D})  = L_{grad}(\overline{D},\hat{D}+\tilde{D}) + L_{SSIM}(\overline{D},\hat{D}+\tilde{D}) \eqno{(3)}
$$
$$
\setlength{\abovedisplayskip}{3pt}
\setlength{\belowdisplayskip}{3pt}
L_{grad}(\overline{D},\hat{D}+\tilde{D}) = \frac{1}{n} \sum_{i=1}^n |\nabla_x (\overline{D_i},\hat{D_i}+\tilde{D_i})| +  | \nabla_y (\overline{D_i},\hat{D_i}+\tilde{D_i}) | \eqno{(4)}
$$
$$
\setlength{\abovedisplayskip}{3pt}
\setlength{\belowdisplayskip}{3pt}
L_{SSIM}(\overline{D},\hat{D}+\tilde{D}) = \frac{1}{2} ( 1 - SSIM(\overline{D},\hat{D}+\tilde{D})) \eqno{(5)}
$$

We simply set $\lambda = 1$ in Eq (\ref{eq_loss}) throughout the experiments.

\begin{figure*}[t]
  \centering
  \subfloat[RMSE\_GT+]{
    \centering
    \label{fig4:gt+}
    \includegraphics[width=0.2\textwidth]{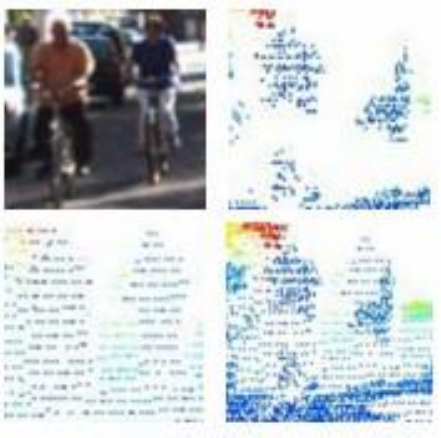} }
  \subfloat[Edge areas used by RMSE\_Edge]{
    \centering
    \label{fig4:edge}
    \includegraphics[width=0.6\textwidth]{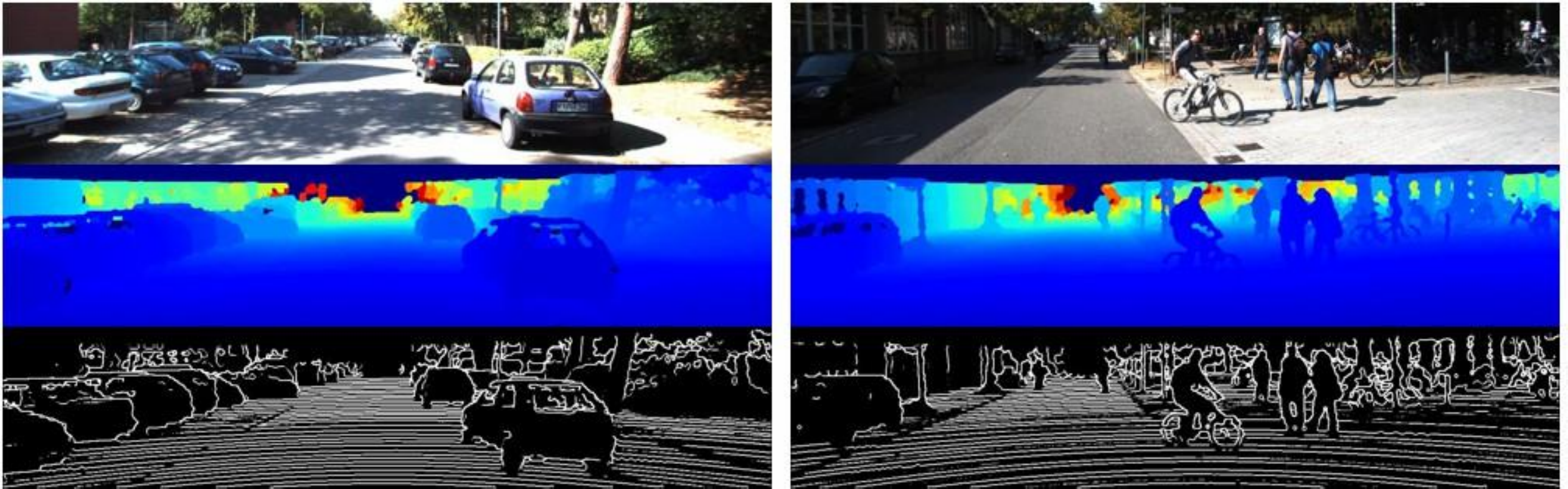} }
  \caption{\textbf{Illustration of GT+ and high gradient areas in pseudo depth map. }  In (a), From top to bottom, left to right: the RGB crop of two cyclists, the corresponding crop in GT, raw sparse depth and our complemented GT+. In (b), two examples of edge areas used by RMSE\_Edge. From top to bottom: RGB image, pseudo ground truth depth and high-gradient areas in the pseudo depth map (marked white).}
  \label{fig4:newmetrics}
\end{figure*}

\subsection{New Evaluation Metrics}

While RMSE and MAE are effective metrics for evaluating dense depth output, it cannot well represent the true quality of quality in terms of 3D shape and structure, as previously illustrated in Fig \ref{fig1:pointcloud}. This is mainly due to the following two reasons: (1) The excessive deletion of pixels in small and moving objects when constructing the ground truth; (2) The lack of quality evaluation on object boundaries. For each scan of sparse LiDAR in KITTI dataset\cite{uhrig2017sparsity}, the ground truth for depth completion is generated by accumulating multiple sweeps of LiDAR scans and projecting to the image space. Most of the possible mixed-depth areas are simply removed instead of rectified\cite{uhrig2017sparsity}. In order to remove outliers caused by occlusion, dynamic objects or measurement artifacts, most points in small or moving objects influenced by the accumulation are further removed. This results in a quasi-dense ground truth focusing more on ground and static backgrounds. Better metrics capable of representing the structure and shape details of the scene are highly demanded.

\textbf{RMSE\_GT+.} We seek to make a supplement for ground truth with the rectified sparse depth from current frame. Rectified sparse is free from penetration problem and still contains many measurement points in small or moving objects. This supplemented ground truth is called Ground Truth+ (GT+). It contains more effective pixels than the original one, as shown in Fig \ref{fig4:gt+}. We also use root mean square error between the prediction and GT+ namely RMSE\_GT+ to evaluate the quality of the scene structure.

\textbf{RMSE\_Edge.} As illustrated in Fig \ref{fig1:deeplidar1}(e), RMSE over the entire image cannot well represent the prediction quality on object boundaries. To focus more on edge areas of depth map, we can set up a metric evaluating the accuracy only on those edge pixels. We set the mean value of the gradient of pseudo depth map as a threshold to locate the edge pixels, as shown in Fig \ref{fig4:edge}. RMSE\_Edge metric is defined as the RMSE error between the predicted depth and GT+ on those edge pixels. Note that both RMSE\_GT+ and RMSE\_Edge metrics are only used for evaluation purpose.

\section{Experiments}

Extensive experiments on public dataset are conducted to verify the effectiveness of our method, as well as the validity of the proposed new metrics.

\subsection{Setup}

We evaluate our model on publicly available KITTI dataset. KITTI depth completion dataset\cite{uhrig2017sparsity} contains 86,898 frames for training, 1,000 frames for validation and another 1,000 frames for testing. During training, we ignored regions without LiDAR points (\textit{i}.\textit{e}. top 100 rows) and center crop 512x256 patches as training examples. We train the network for 25 epochs on two NVIDIA 2080Ti GPUs with batch size of 12. We use the Adam optimizer\cite{2014Adam} and set the initial learning rate 0.001, which is decayed by 0.5 every 5 epochs after the first 10 epochs. The network is trained from scratch without inducing any additional data.

\subsection{Comparative Results}

\begin{table*}[t]
  \caption{\textbf{Performance comparsion on KITTI Benchmark}(Unit in \emph{mm}).}
  \label{table:test}
  \begin{center}
  \scalebox{1.0}{
  \begin{tabular}{c|c|c|c|c|c|c}
  \toprule
  Method & Realtime &RMSE & MAE & iRMSE & iMAE & FPS\\
  \midrule
  B-ADT \cite{9130033} & No & 1480.36 & 298.72 & 4.16 & 1.23 & 8.3 \\
  CSPN\cite{cheng2018depth}  & No & 1019.64 & 279.46 & 2.93 & 1.15 & 1 \\
  DC\_coef\cite{depth-coefficients-for-depth-completion}  & No&965.87 & 215.75 & 2.43 & 0.98 & 6.7  \\
  SSGP \cite{schuster2021ssgp} & No & 838.22 & 244.70 & 2.51 & 1.09 & 7.1\\
  CG\_Net \cite{lee2020deep} & No & 807.42 & 253.98 & 2.73 & 1.33 & 5\\
  CSPN++\cite{CSPNplus}  & No&743.69 & 209.28 & 2.07 & 0.90 & 5 \\
  NLSPN\cite{park2020non}   &No &\bf 741.68 & \bf 199.59 & \bf 1.99 & \bf 0.84 & 4.5  \\
  \midrule
  pNCNN\cite{Eldesokey_2020_CVPR}  & Yes &960.05 & 251.77 & 3.37 & 1.05 & \bf 50 \\
  Sparse to Dense\cite{ma2019self}  & Yes & 954.36 & 288.64 & 3.21 & 1.35 & 25  \\
  PwP\cite{yan2019completion}  & Yes &777.05 & 235.17 & 2.42 & 1.13 & 10  \\
  DeepLiDAR\cite{qiu2019deeplidar}  & Yes &758.38  & 226.50 & 2.56  & 1.15 & 14.3 \\
  UberATG\cite{learning2019yun}  & Yes & \bf 752.88 & 221.19 & 2.34 & 1.14 & 11.1 \\
  \midrule
  Ours  & Yes & 755.41 & \bf 214.13 & \bf2.25 & \bf0.96 & \bf 50 \\
  \bottomrule
  \end{tabular}}
  \end{center}
\end{table*}

Table \ref{table:test} shows the comparative results of our model on KITTI depth completion benchmark, where the evaluation is carried out on KITTI testing server automatically. Our method achieves state-of-the-art performance with the highest frame rate at 50 Hz. Except for RMSE which is very close to \cite{learning2019yun}, other metrics of our method rank the first among all real-time methods. In comparison to the latest non-real-time works, \textit{e}.\textit{g}., CSPN++\cite{CSPNplus} and NLSPN\cite{park2020non}, our model achieves close performance, but runs 10 times faster, which creates more possibilities for downstream tasks.

\subsection{Evaluation on New Metrics}

Before using the new metrics, we first verify the quality and the credibility of the rectified sparse depth and the complemented ground truth (GT+). Towards this goal, we follow the process of creating the ground truth of KITTI depth completion dataset\cite{uhrig2017sparsity}, which exploits the manually cleaned KITTI 2015 stereo disparity maps\cite{Menze2018JPRS} as reference. The disparity maps are transformed into depth values using the calibration files provided by KITTI. The evaluation results are listed in Table \ref{table:2015}, where "KITTI outliers" is defined as the ratio of pixels whose depth value larger than 3 meters and the relative depth error larger than 5\%, just as \cite{uhrig2017sparsity}.

\begin{table}[t]
\caption{\textbf{Evaluation of Depth Maps Using KITTI 2015 Training Dataset.} All metrics are in \emph{mm}.}
\label{table:2015}
\begin{center}
\scalebox{1.0}{
\begin{tabular}{c|c|c|c|c}
\toprule
Method & Density & MAE & RMSE & KITTI Outliers\\
\midrule
Sparse LiDAR  & 3.99\% & 509.04 & 2544.62 & 2.12\% \\
Rectifed Sparse  & 3.54\% & \bf 255.66 & \bf 643.66 & \bf 0.87\% \\
\midrule
GroundTruth\cite{uhrig2017sparsity}  & 14.43\% & 388.28 & 938.00 & 2.33\% \\
GroundTruth+  & 15.80\% & \bf 371.92 & \bf 923.12 & \bf 2.21\% \\
\bottomrule
\end{tabular}}
\end{center}
\end{table}

From Table \ref{table:2015}, where we can see both rectified sparse and GT+ have better quality than their counterpart in all three metrics (MAE, RMSE and KITTI Outliers). For the rectified sparse depth, after removing those mixed-depth pixels with the help of pseudo depth map, most points are reserved with significant quality improvement. For GT+, which fuses the rectified sparse into the original ground truth, it contains about 1.50\% more high-confidence points and less outliers than the original ground truth. Therefore, GT+ are convincible for evaluating the quality of the dense prediction.

\subsection{Ablation Studies}

In order to investigate the impact of different modules on the final result, we conduct ablation studies on the KITTI validation dataset. The results are shown in Table \ref{table:module}, where Baseline, Rectification, Structural loss and Residual represent backbone with a single DCU, rectification on sparse depth, training with structural loss and predicting the residual depth, respectively.

\begin{table}[t]
  \caption{\textbf{Ablation Study on KITTI Validation Set.(mm) }}
  \label{table:module}
  \begin{center}
  \scalebox{0.7}{
  \begin{tabular}{c|c|c|c|c|c|c}
  \toprule
  Baseline & Rectification & StructuralLoss & Residual & RMSE & RMSE\_GT+ & RMSE\_Edge \\
  \midrule
  \checkmark & & & & 829.87 & 1596.22 & 2794.98 \\
  \checkmark & \checkmark & &  & 822.10 & 1572.31 & 2367.21  \\
  \checkmark & \checkmark &\checkmark &  & 810.38 & 1392.73 & 2271.23  \\
  \checkmark & \checkmark& \checkmark& \checkmark & \bf 795.97 & \bf 1335.20 & \bf 2171.76 \\
  \bottomrule
  \end{tabular}}
  \end{center}
\end{table}

As shown in Table \ref{table:module}, adding each design leads to a performance rise. The final complete model behaves the best upon all of the three metrics. Comparing with the baseline, our final model has performance improvement at about 4.08\% in RMSE, 16.40\% in RMSE\_GT+ and 22.30\% in RMSE\_Edge, respectively.

\begin{figure}[t]
  \centering
  \subfloat[Before Post-processing]{
    \centering
    \label{fig7:withoutpp}
    \includegraphics[width=0.2\textwidth]{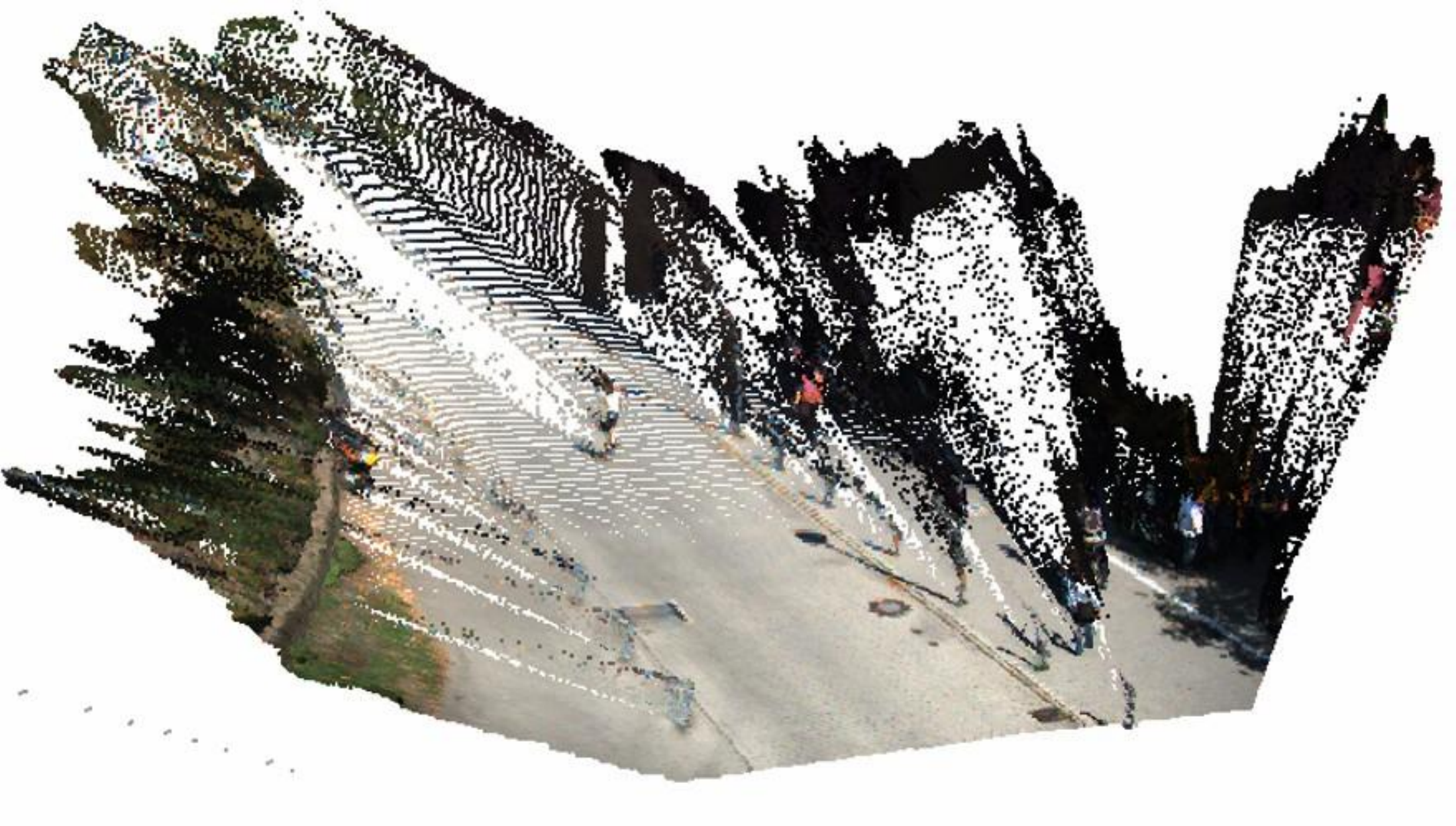} }
  \subfloat[After Post-processing]{
    \centering
    \label{fig7:withpp}
    \includegraphics[width=0.2\textwidth]{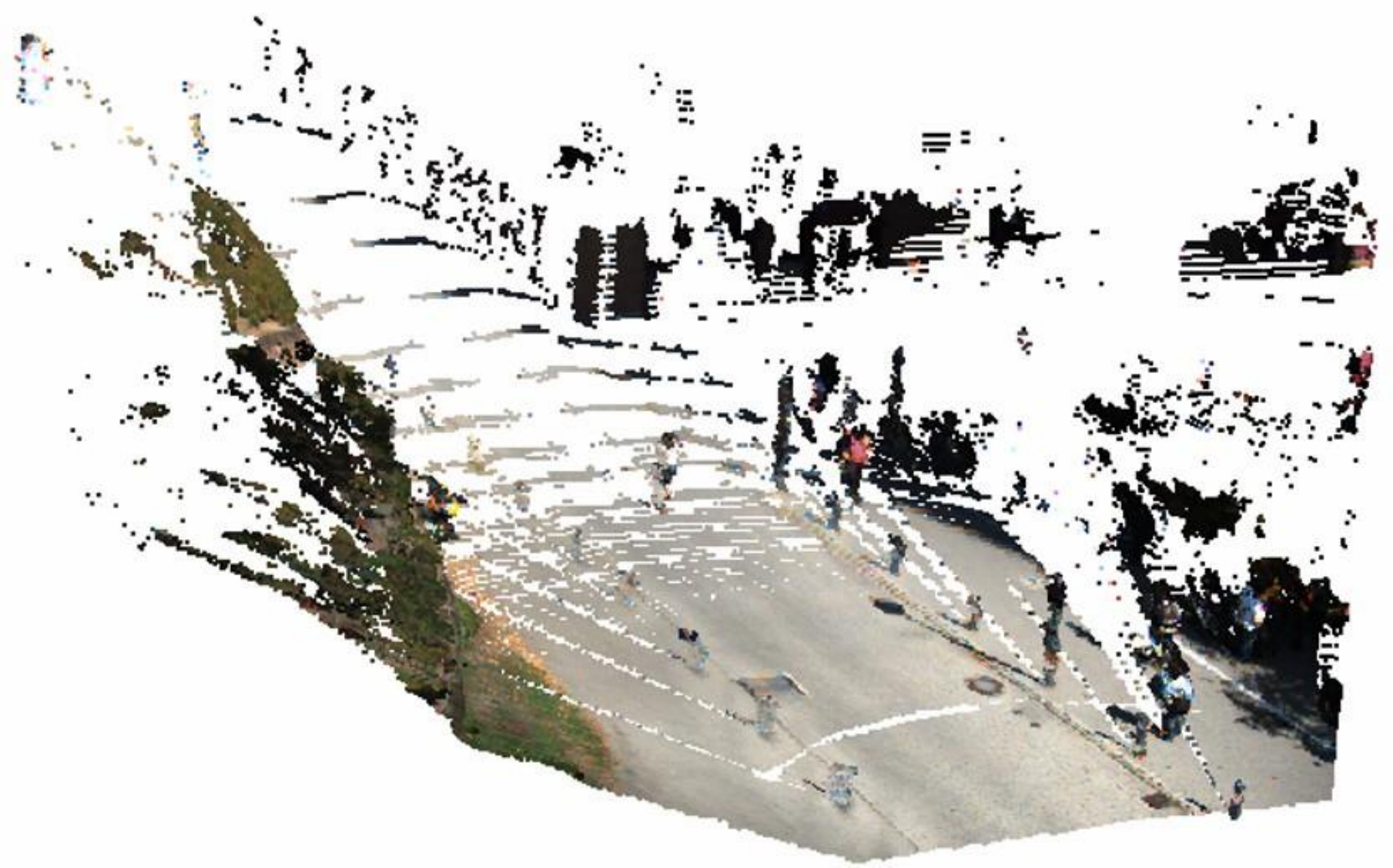} }
  \caption{\textbf{Illustration of dense point cloud before/after post-processing.}  }
  \label{fig7:pp}
\end{figure}

\begin{figure*}[t]
  \centering
  \subfloat{
    \centering
    \includegraphics[width=0.3\textwidth]{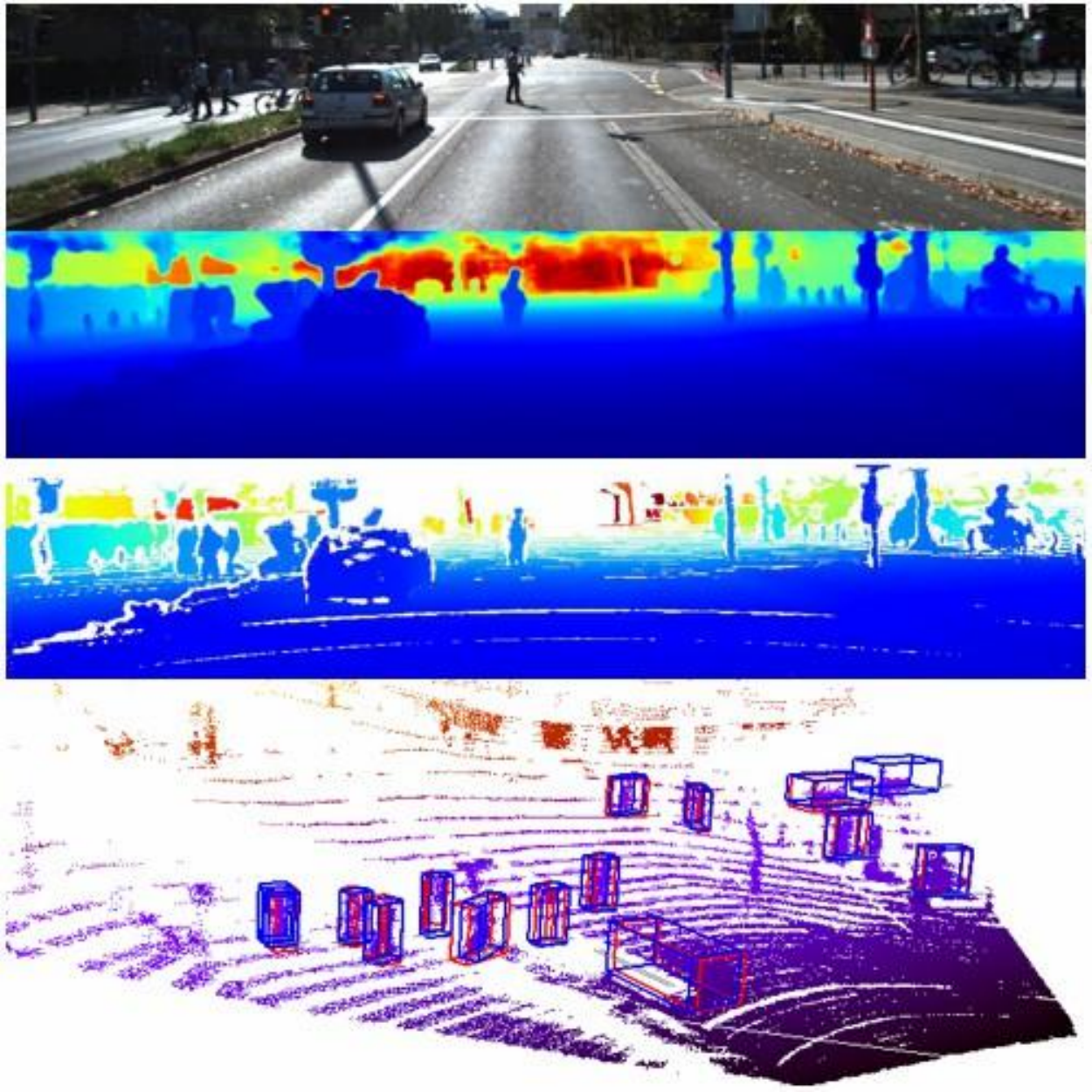} }
  \subfloat{
    \centering
    \includegraphics[width=0.3\textwidth]{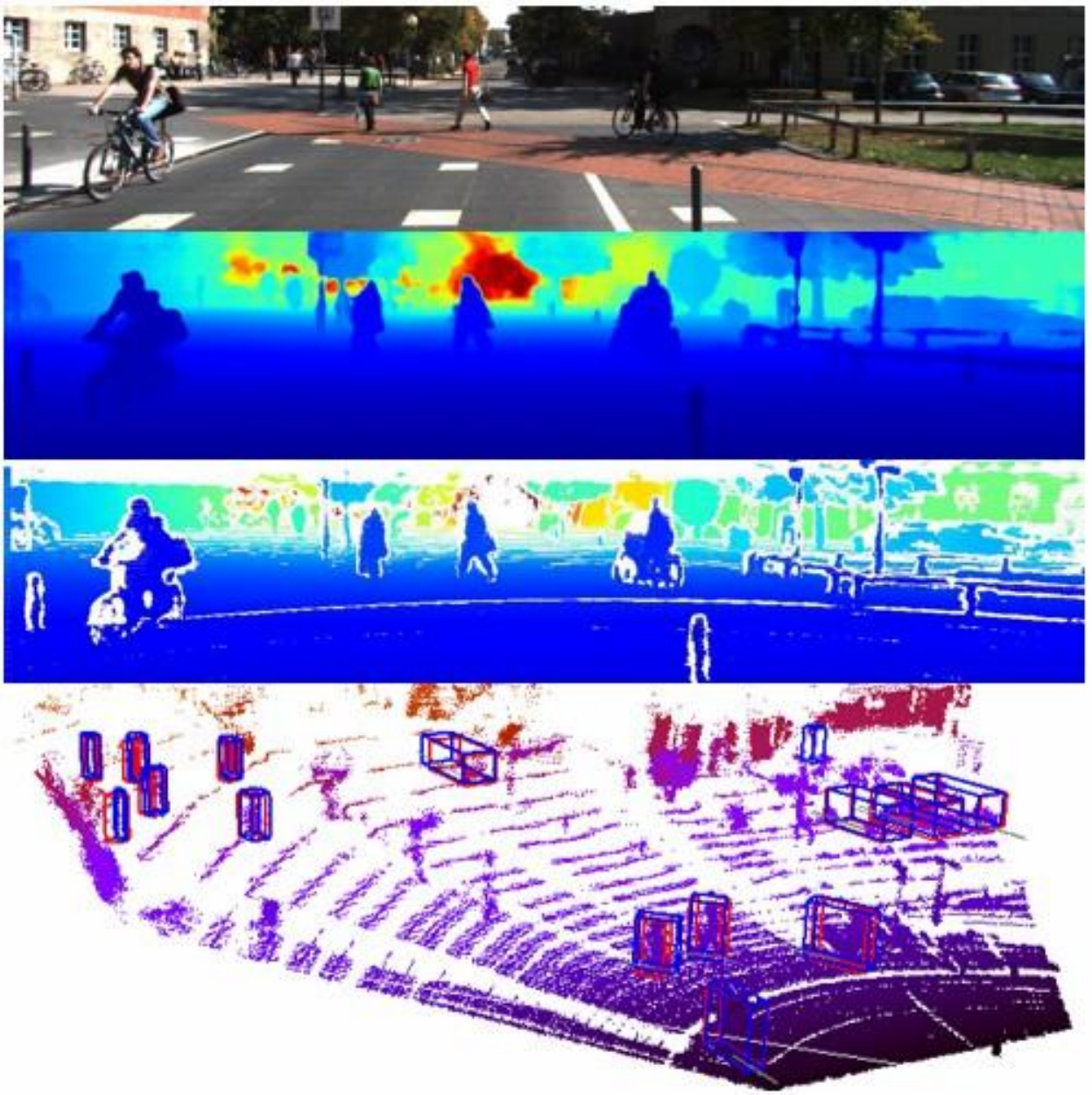} }
    \subfloat{
    \centering
    \includegraphics[width=0.3\textwidth]{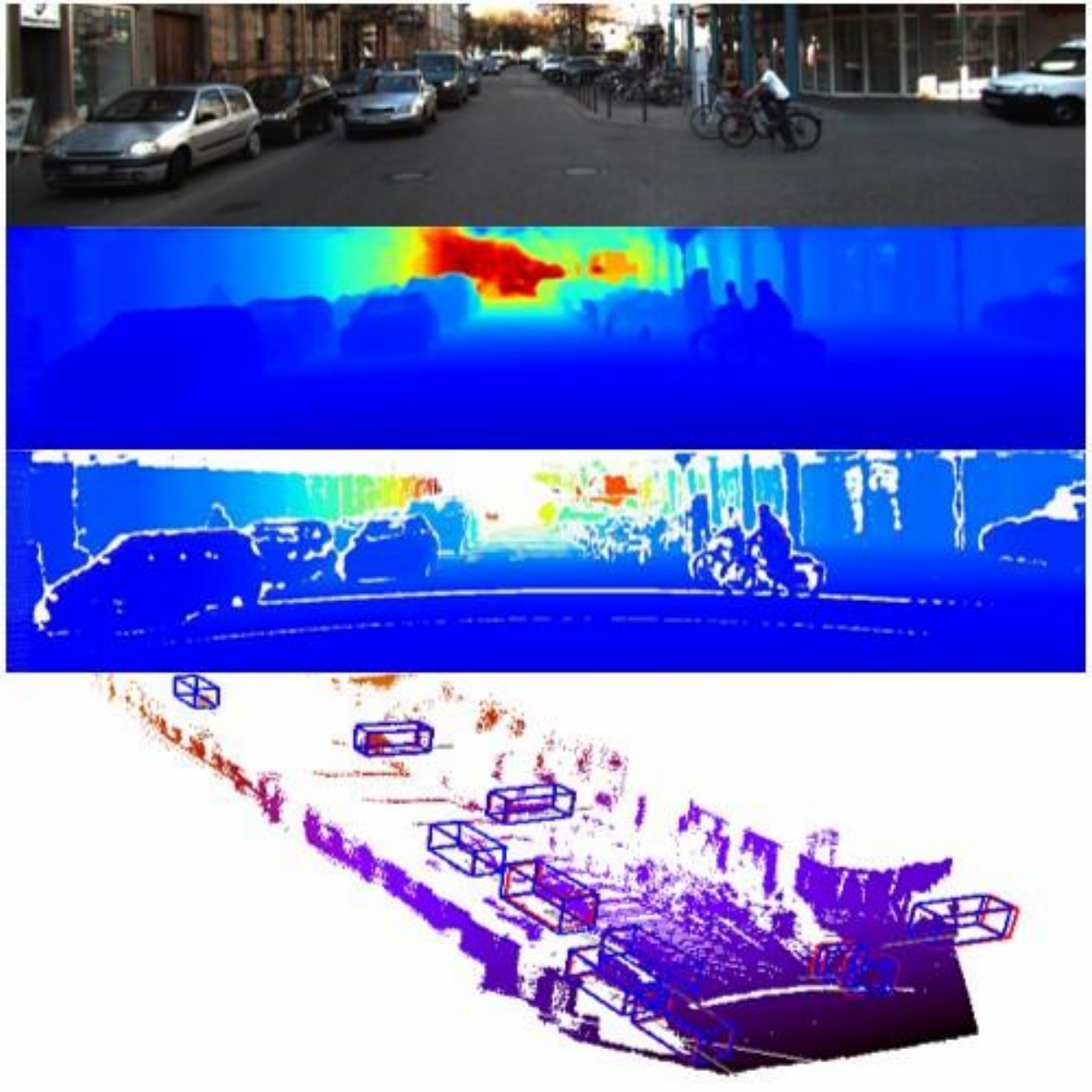} }

  \caption{\textbf{Qualitative results on KITTI 3D Object Detection Dataset. } This figure lists RGB image, predicted dense depth before and after post-processing and 3D object detection results. GroundTruth and predictions are marked by blue and red boxes, respectively.}
  \label{fig6:3ddetection}
\end{figure*}

\subsection{Generalization Studies}
\subsubsection{Sparsity variation}

To verify the generalization of our method on LiDAR data with different degree of sparsity, we conduct the ablation experiments on LiDAR scans with 64, 32 and 16 lines, respectively. Due to the lack of depth completion datasets produced on real 32 and 16-line LiDARs, we synthesize them by subsampling the KITTI 64-line LiDAR data. To keep the same scanning manner as real Lidars, the subsampling is carried out on a range-view image with elevation and azimuth angles as two coordinate axises. The obtained downsampled 3D points are then projected to RGB image plane to produce sparse depth images. We retrain the network for each data modal before evaluation and the error of the completed depth are listed in Table \ref{table:sparsedepth}. 

\begin{table}[t]
  \caption{\textbf{Sparisty Experiments.(mm)}}
  \label{table:sparsedepth}
  \begin{center}
  \scalebox{0.9}{
  \begin{tabular}{c|c|c|c|c} 
    \toprule
    LiDAR resolution & RMSE & MAE & RMSE\_GT+ & RMSE\_Edge\\ 
    \midrule
    64 lines(Baseline) & 829.87 & 240.32 & 1596.22 & 2794.98 \\
    64 lines(Ours) & 795.97 & 213.64  & 1335.20 & 2171.76 \\
    32 lines(Baseline) & 2031.23 & 597.12  & 3494.53 & 5153.33 \\
    32 lines(Ours) & 1780.55 & 540.28  & 2833.96 & 4348.86 \\
    16 lines(Baseline) & 2357.61 & 695.28 & 4072.46 & 5743.28 \\
    16 lines(Ours) & 1989.48 & 592.78  & 3256.02 & 4588.20  \\ 
    \bottomrule
  \end{tabular}}
  \end{center}
\end{table}

For all of the data modals our DenseLiDAR performs much better than the baseline method, and the leading gaps are getting larger with respect to the sparser depth input. On the other hand, sparser depth input does bring challenge to the completion task. As shown in Table \ref{table:sparsedepth},  depth input of 32-line results in about three times larger errors in all metrics than those of 64-line data.  Feeding depth of 16-line further expands the gap. We can infer that with the reduction of LiDAR’s scan lines, the network has to rely more on RGB image features and becomes much closer to a pure depth prediction work.

\subsubsection{Indoor dataset}
\begin{table}[t]
  \caption{\textbf{RMSE Metric Performances in NYUv2 Dataset.(m)}}
  \label{table:nyuv2}
  \begin{center}
  \scalebox{0.8}{
  \begin{tabular}{c|c|c|c|c|c|c|c} 
    \toprule
    Sampled Points & 100 & 300 & 500 & 1000 & 3000 & 5000 & 10000\\ 
    \midrule
    DeepLiDAR\cite{qiu2019deeplidar} & 0.381 & 0.339 & 0.320 & 0.289 & 0.255 & 0.189 & 0.149\\
    Ours & 0.298 & 0.270 & 0.258 & 0.231 & 0.199 & 0.147 & 0.101\\
    \bottomrule
  \end{tabular}}
  \end{center}
\end{table}

We conduct experiments on NYUv2 dataset\cite{Silberman:ECCV12}, which consists of dense RGBD images collected from 464 indoor scenes and has very different depth distribution patterns to Lidar. We use a uniform sparsifier with varying sampling numnber to produce sparse depth images from the dense ones. For convenience, we only use the labelled data (about 2K images) in the dataset with a split ratio of 2:1 for training and testing, respectively. The results are shown in Table \ref{table:nyuv2}, where we can see that in all levels of sparsity our method  obtains better results than DeepLiDAR\cite{qiu2019deeplidar}.

\section{Applications}

To evaluate the depth quality and explore the possibility of utilizing the obtained high-quality dense depth maps, we also apply our completed depth in two different robotic tasks: 3D object detection and RGB-D SLAM. To further improve the quality of the resulting point cloud, post-processing can be carried out before taking the predicted dense depth as input. 

\subsection{Post-processing} We utilize the pseudo depth map once again in this step. We compute pixel-wise difference between the predicted depth and its corresponding pseudo depth map. The pixels whose depth difference with the pseudo map larger than a threshold are regarded as outliers and hence removed from the point cloud. In this step, the selection of a reasonable depth threshold becomes essential. We conduct experiments in different thresholds and results are listed in Table \ref{table:postprocessing}. Overall, the quality of the depth image gets to be better when smaller depth threshold is used, but at a cost of density decrease. In particular, RMSE can be halved from 802mm to 426mm when a global threshold of 1 meter is used, with a cost that less than 4\% of points with ground truth are removed. Taking both accuracy and density into account, we set a dynamic threshold based on a piecewise depth range to remove outliers. We set threshold of 0.1m for pixels closer than 10m, 0.3m for pixels between 10m to 40m, and 0.5m for those beyond 40m, respectively. The final RMSE can be further decreased to 324mm with about 89\% points with ground truth retained. An example of the obtained 3D point cloud before and after post-processing is illustrated in Fig \ref{fig7:pp}.

\begin{table}[t]
  \caption{\textbf{Influence of Thresholds Used in Post-processing. }}
  \label{table:postprocessing}
  \begin{center}
  \scalebox{0.7}{
  \begin{tabular}{c|c|c|c|c|c|c}
  \toprule
  Threshold & None & 10m & 5m & 3m & 1m & Dynamic \\
  \midrule
  Remained GT Points  & 100.00\% & 99.67\% & 99.28\% & 98.81\% & 96.63\% & 89.70\% \\
  \midrule
  RMSE(/mm)  & 802.16  & 672.12 & 605.38 & 551.84 & 426.19 & 324.50  \\
  \bottomrule
  \end{tabular}}
  \end{center}
\end{table}

\subsection{3D Object Detection} 

We choose F-PointNet\cite{qi2018frustum} as our object detection method to verify the quality of the produced dense depth. F-PointNet\cite{qi2018frustum} projects depth pixels inside a 2D bounding box into 3D point cloud and performs object detection within this point frustum, where the sampling points in a single frustum is fixed to 1,024. For most distant or small objects, the number of measurement point can fall down to 50~200 points or even less, which means a lot of points have to be sampled repeatedly to meet the constraint of 1,024. More effective dense points can possibly contribute to better detection results. In our experiments, we first sample points from original sparse data, then if the point number within the frustum is less than 1,024, we add the predicted dense points from depth completion. We follow the same training and validation splits as \cite{chen20153d} in KITTI 3D object detection dataset\cite{Geiger2012CVPR}, which contains 3712 samples for training and 3769 for testing respectively.

\begin{table}[h]
  \caption{\textbf{$AP_{3D}$(in \%) of 3D Object Detection on KITTI Dataset.} PP Stands for Post-processing. }
  \label{table:3ddetection}
  \begin{center}
  \scalebox{0.7}{
  \begin{tabular}{c|c|c|c|c|c|c|c|c|c}
  \toprule
  \multirow{2}{*}{Method}  & \multicolumn{3}{|c|}{$Car$}  &\multicolumn{3}{|c|}{$Pedestrian$} & \multicolumn{3}{|c}{$Cyclist$}\\
  \cmidrule{2-10}
  & Eas. & Mod. & Har. &  Eas. & Mod. & Har. &  Eas. & Mod. & Har.\\
  \midrule
  Sparse LiDAR  & 83.83 & 71.17 & 63.28 & 64.93 & 55.61 & 49.14 & 76.20 &  56.91 & 53.07 \\
  DeepLiDAR  & 78.99 & 62.13 & 53.93 & 63.98 & 54.05 & 46.73 & 62.42 & 44.72 & 41.53\\
  DCU  & 78.43 & 61.43  & 51.39 & 59.95 & 50.74 & 43.15 & 62.84 & 43.39 & 41.18\\
  \midrule
  Ours(w/o PP)  & \bf 84.47 & 68.92 & 61.21 & \bf 67.22 & \bf 56.50 & 48.92 & \bf 79.95 & 55.75 & 51.43\\
  Ours(w PP) & \bf 85.52 & \bf 72.33 & \bf 64.39 & \bf 69.88 & \bf 60.88 & \bf 52.36 & \bf 83.70 & \bf 60.16 & \bf 56.02\\
  \bottomrule
  \end{tabular}}
  \end{center}
\end{table}

\begin{table*}[!t]
  \caption{\textbf{Quantative Results of KITTI Odometry Sequences.}'-' Denotes Tracking Failure. 'MP' Denotes Average Matched ORB Points}
  \label{table:rgbdslam}
  \begin{center}
  \scalebox{0.8}{
  \begin{tabular}{c|c|c|c|c|c|c|c|c|c|c|c|c}
  \toprule
  $t_{rel}(\%)$ / $r_{rel}(deg/100m)$          & 00       &  01 & 02  & 03 & 04  & 05  &   06     & 07       & 08    & 09    & 10   & MP  \\
  \midrule
  Monocular & 0.80/1.32 &  - & - &  \textbf{0.49}/0.37  &   0.70/0.25  &  0.88/0.90   & 1.10/0.42    & 0.89/0.23    & 3.06/1.41 & -   & 1.01/9.16 &  162\\
  Stereo            & 0.71/0.25   & \textbf{1.48}/\textbf{0.21} & 0.80/0.26 & 0.80/\textbf{0.20} & \textbf{0.47}/\textbf{0.15} & \textbf{0.39}/\textbf{0.16} &  0.47/\textbf{0.15}   & 0.49/0.28 & \textbf{1.03}/\textbf{0.30} & 0.89/0.26  & \textbf{0.66}/0.30  &  318\\
  +SparseD    &  -   & -   &  -  & -  & -  & -   & -   & -    & - & - & - &  0 \\
  +DeepLiDARD    & 0.83/0.39  &  7.41/2.43  & 1.00/0.37   & 1.25/0.39 & 3.83/3.18 & 1.30/0.36 & 3.52/1.08 & 1.08/0.45  & 2.23/0.60 & 2.44/0.47 & 2.83/1.01 &  359 \\
  \midrule
  +OursD(w/o PP) & 0.81/0.38 & 45.70/9.24 & 1.03/0.39   & 1.24/0.37 & 1.17/1.29 & 0.51/0.32 & 0.57/0.39    & 0.46/0.38    & 1.33/0.48 &  0.94/0.35  & 0.89/0.53  &  417\\
  +OursD(w PP)   & \textbf{0.70}/\textbf{0.24}  & 19.71/6.21 & \textbf{0.77}/\textbf{0.25}  & 1.13/0.23 & 0.71/0.47 & 0.53/0.22 & \textbf{0.43}/0.19 & \textbf{0.44}/\textbf{0.25} & 1.04/0.33 & \textbf{0.85}/\textbf{0.25}  & 0.94/\textbf{0.29}  &  402 \\
  \bottomrule
  \end{tabular}}
  \end{center}
\end{table*}

We report $AP_{3D}$(in\%) which corresponds to average precision with 40 recall positions of 3D bounding box with rotated IoU threshold 0.7 for Cars and 0.5 for Pedestrian and Cyclist, respectively. The evaluation results are listed in Table \ref{table:3ddetection}. Instead of achieving better results, directly using the dense depth from DeepLiDAR\cite{qiu2019deeplidar} or baseline (a single DCU backbone) produces much worse performance than using the raw sparse data. This is due to the distorted object shape caused by the erroneous mixed-depth points around object boundaries, as shown in Fig \ref{fig1:deeplidar2}. They would have a severer negative impact on regression and classification of objects. On the other hand, using the high-quality dense depth from our DenseLiDAR leads to much better detection performance than DeepLiDAR and baseline. Using the point cloud after post-processing, the highest AP values are achieved in all three difficulty levels. Comparing with raw sparse data in Table \ref{table:3ddetection}, using our dense point cloud after post-processing achieves about 1.10\% AP increase for $Cars$, 5.20\% for $Pedestrian$ and 3.25\% for $Cyclist$ in Moderate difficulty level, respectively. Thanks to the dense and accurate 3D point clouds from our DenseLiDAR, the geometric features of the objects are highly enhanced, resulting in a notable performance improvement. The result also shows that small or distant objects could benefit more from denser point clouds than large objects. Some qualitative 3D object detection results using our dense point cloud are also illustrated in Fig \ref{fig6:3ddetection}. 

\subsection{RGB-D SLAM}

We choose open-source ORB-SLAM2 \cite{murORB2}, a popular real-time SLAM library for Monocular \cite{murTRO2015}, Stereo or RGB-D camera configurations, as our evaluation baseline. With the dense depth maps, we are able to run the RGB-D mode of ORB-SLAM2 in KITTI odometry dataset. For all of the 11 sequences, we run each of them for 25 times and record the average relative translation $t_{rel}$ and rotation $r_{rel}$ errors in Table \ref{table:rgbdslam}.

We have some interesting findings in Table \ref{table:rgbdslam}. Firstly, RGB image with a sparse depth map is not applicable for RGB-D mode, where depth value for every ORB feature point is required. Using the dense completed depth, good positioning results can be obtained with RGB-D mode. Secondly, except for sequence 01, our performance is robust across different sequences and much better than using depths from other depth completion methods or pure monocular images. 
Sequence 01 is a challenging highway scene with large area of sky and little structured objects alongside. Few training samples for depth completion are from this kind of scene, which results in dense depth image with large error as input of RGB-D SLAM and leads to some kinds of failure. Thirdly, our method is able to produce more matched feature points on average than other modes because of more reliable depth pixels, as shown in last column of Table \ref{table:rgbdslam}. It can also contribute to better positioning accuracy. Lastly, the post-processing for removing outliers in dense depth is also helpful in RGB-D SLAM. 

\begin{figure}[thpb]
  \centering
  \includegraphics[width=0.45\textwidth]{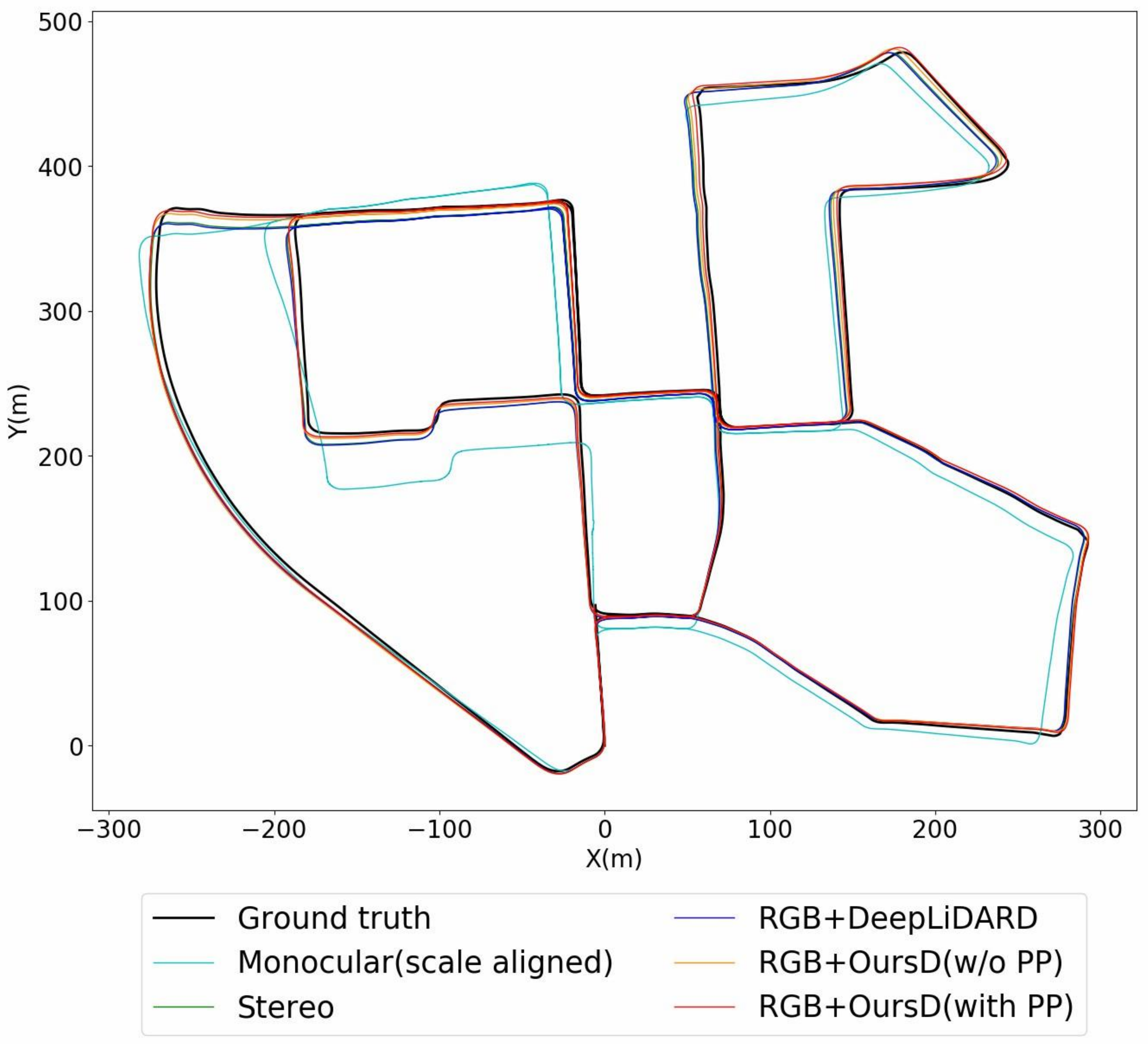} 
  \captionsetup{justification=centering}
  \caption{\textbf{Estimated Trajectory of KITTI 00.}}
  \label{fig:trajectory}
\end{figure}

As an typical example, the positioning results estimated for sequence 00 are illustrated in Fig \ref{fig:trajectory}, where we can observe that the trajectory of our method is closer to the ground truth than other methods.

\section{Conclusions}

We present DenseLiDAR, a novel real-time pseudo depth guided depth completion network. We exploit the pseudo depth map in constructing a residual-based prediction, rectifying sparse input and supervising the network with a structural loss. We point out that popular RMSE metric is not sensitive for evaluating the real quality difference of the completed depth because of its sparsity. We propose RMSE\_GT+ and RMSE\_Edge metrics to better represent the true quality of the prediction. Experimental results on KITTI depth completion dataset demonstrate that our model achieves comparable performance to the state-of-the-art methods on both RMSE and new metrics with the highest running speed. Extensive experiments in applying the dense depth in 3D object detection and RGB-D SLAM also verify the quality improvement of our prediction and further demonstrate the potential benefits of combining our depth completion with other downstream perception or localization tasks in robotic fields.



\bibliographystyle{IEEEtran}
\bibliography{IEEEabrv,ref}

\begin{thebibliography}{10}
\providecommand{\url}[1]{#1}
\csname url@rmstyle\endcsname
\providecommand{\newblock}{\relax}
\providecommand{\bibinfo}[2]{#2}
\providecommand\BIBentrySTDinterwordspacing{\spaceskip=0pt\relax}
\providecommand\BIBentryALTinterwordstretchfactor{4}
\providecommand\BIBentryALTinterwordspacing{\spaceskip=\fontdimen2\font plus
\BIBentryALTinterwordstretchfactor\fontdimen3\font minus
  \fontdimen4\font\relax}
\providecommand\BIBforeignlanguage[2]{{%
\expandafter\ifx\csname l@#1\endcsname\relax
\typeout{** WARNING: IEEEtran.bst: No hyphenation pattern has been}%
\typeout{** loaded for the language `#1'. Using the pattern for}%
\typeout{** the default language instead.}%
\else
\language=\csname l@#1\endcsname
\fi
#2}}

\bibitem{qiu2019deeplidar}
J.~Qiu, Z.~Cui, Y.~Zhang, X.~Zhang, S.~Liu, B.~Zeng, and M.~Pollefeys,
  ``Deeplidar: Deep surface normal guided depth prediction for outdoor scene
  from sparse lidar data and single color image,'' in \emph{Proceedings of the
  IEEE Conference on Computer Vision and Pattern Recognition}, 2019, pp.
  3313--3322.

\bibitem{uhrig2017sparsity}
J.~Uhrig, N.~Schneider, L.~Schneider, U.~Franke, T.~Brox, and A.~Geiger,
  ``Sparsity invariant cnns,'' in \emph{2017 international conference on 3D
  Vision (3DV)}.\hskip 1em plus 0.5em minus 0.4em\relax IEEE, 2017, pp. 11--20.

\bibitem{hawe2011dense}
S.~Hawe, M.~Kleinsteuber, and K.~Diepold, ``Dense disparity maps from sparse
  disparity measurements,'' in \emph{2011 International Conference on Computer
  Vision}.\hskip 1em plus 0.5em minus 0.4em\relax IEEE, 2011, pp. 2126--2133.

\bibitem{liu2015depth}
L.-K. Liu, S.~H. Chan, and T.~Q. Nguyen, ``Depth reconstruction from sparse
  samples: Representation, algorithm, and sampling,'' \emph{IEEE Transactions
  on Image Processing}, vol.~24, no.~6, pp. 1983--1996, 2015.

\bibitem{ku2018defense}
J.~Ku, A.~Harakeh, and S.~L. Waslander, ``In defense of classical image
  processing: Fast depth completion on the cpu,'' in \emph{2018 15th Conference
  on Computer and Robot Vision (CRV)}.\hskip 1em plus 0.5em minus 0.4em\relax
  IEEE, 2018, pp. 16--22.

\bibitem{6618993}
M.~{Hornácek}, C.~{Rhemann}, M.~{Gelautz}, and C.~{Rother}, ``Depth super
  resolution by rigid body self-similarity in 3d,'' in \emph{2013 IEEE
  Conference on Computer Vision and Pattern Recognition}, 2013, pp. 1123--1130.

\bibitem{ma2019self}
F.~Ma, G.~V. Cavalheiro, and S.~Karaman, ``Self-supervised sparse-to-dense:
  Self-supervised depth completion from lidar and monocular camera,'' in
  \emph{2019 International Conference on Robotics and Automation (ICRA)}.\hskip
  1em plus 0.5em minus 0.4em\relax IEEE, 2019, pp. 3288--3295.

\bibitem{zhong2019deep}
Y.~Zhong, C.-Y. Wu, S.~You, and U.~Neumann, ``Deep rgb-d canonical correlation
  analysis for sparse depth completion,'' in \emph{Advances in Neural
  Information Processing Systems}, 2019, pp. 5332--5342.

\bibitem{lee2020deep}
S.~Lee, J.~Lee, D.~Kim, and J.~Kim, ``Deep architecture with cross guidance
  between single image and sparse lidar data for depth completion,'' \emph{IEEE
  Access}, 2020.

\bibitem{cheng2018depth}
X.~Cheng, P.~Wang, and R.~Yang, ``Depth estimation via affinity learned with
  convolutional spatial propagation network,'' in \emph{Proceedings of the
  European Conference on Computer Vision (ECCV)}, 2018, pp. 103--119.

\bibitem{CSPNplus}
X.~Cheng, P.~Wang, G.~Chenye, and R.~Yang, ``Cspn++: Learning context and
  resource aware convolutional spatial propagation networks for depth
  completion,'' \emph{Proceedings of the AAAI Conference on Artificial
  Intelligence}, vol.~34, pp. 10\,615--10\,622, 04 2020.

\bibitem{xu2020deformable}
Z.~Xu, Y.~Wang, and J.~Yao, ``Deformable spatial propagation network for depth
  completion,'' 2020.

\bibitem{park2020non}
J.~Park, K.~Joo, Z.~Hu, C.-K. Liu, and I.-S. Kweon, ``Non-local spatial
  propagation network for depth completion,'' in \emph{European Conference on
  Computer Vision, ECCV 2020}.\hskip 1em plus 0.5em minus 0.4em\relax European
  Conference on Computer Vision, 2020.

\bibitem{learning2019yun}
Y.~Chen, B.~Yang, M.~Liang, and R.~Urtasun, ``Learning joint 2d-3d
  representations for depth completion,'' in \emph{ICCV}, 2019.

\bibitem{hekmatian2019conf}
H.~Hekmatian, S.~Al-Stouhi, and J.~Jin, ``Conf-net: Predicting depth completion
  error-map for high-confidence dense 3d point-cloud,'' \emph{ArXiv}, vol.
  abs/1907.10148, 2019.

\bibitem{depth-coefficients-for-depth-completion}
S.~Imran, Y.~Long, X.~Liu, and D.~Morris, ``Depth coefficients for depth
  completion,'' in \emph{In Proceeding of IEEE Computer Vision and Pattern
  Recognition}, Long Beach, CA, January 2019.

\bibitem{yan2019completion}
Y.~Xu, X.~Zhu, J.~Shi, G.~Zhang, H.~Bao, and H.~Li, ``Depth completion from
  sparse lidar data with depth-normal constraints,'' in \emph{Proceedings of
  the IEEE International Conference on Computer Vision}.\hskip 1em plus 0.5em
  minus 0.4em\relax IEEE, 2019.

\bibitem{van2019sparse}
W.~Van~Gansbeke, D.~Neven, B.~De~Brabandere, and L.~Van~Gool, ``Sparse and
  noisy lidar completion with rgb guidance and uncertainty,'' in \emph{2019
  16th International Conference on Machine Vision Applications (MVA)}.\hskip
  1em plus 0.5em minus 0.4em\relax IEEE, 2019, pp. 1--6.

\bibitem{cordts2016cityscapes}
M.~Cordts, M.~Omran, S.~Ramos, T.~Rehfeld, M.~Enzweiler, R.~Benenson,
  U.~Franke, S.~Roth, and B.~Schiele, ``The cityscapes dataset for semantic
  urban scene understanding,'' in \emph{Proceedings of the IEEE conference on
  computer vision and pattern recognition}, 2016, pp. 3213--3223.

\bibitem{zou2020simultaneous}
N.~Zou, Z.~Xiang, Y.~Chen, S.~Chen, and C.~Qiao, ``Simultaneous semantic
  segmentation and depth completion with constraint of boundary,''
  \emph{Sensors}, vol.~20, no.~3, p. 635, 2020.

\bibitem{zou2020rsdcn}
N.~Zou, Z.~Xiang, and Y.~Chen, ``Rsdcn: A road semantic guided sparse depth
  completion network,'' \emph{Neural Processing Letters}, pp. 1--13, 2020.

\bibitem{gaidon2016virtual}
A.~Gaidon, Q.~Wang, Y.~Cabon, and E.~Vig, ``Virtual worlds as proxy for
  multi-object tracking analysis,'' in \emph{Proceedings of the IEEE conference
  on computer vision and pattern recognition}, 2016, pp. 4340--4349.

\bibitem{2017CARLA}
A.~Dosovitskiy, G.~Ros, F.~Codevilla, A.~Lopez, and V.~Koltun, ``Carla: An open
  urban driving simulator,'' in \emph{Conference on robot learning}.\hskip 1em
  plus 0.5em minus 0.4em\relax PMLR, 2017, pp. 1--16.

\bibitem{hu2019revisiting}
J.~Hu, M.~Ozay, Y.~Zhang, and T.~Okatani, ``Revisiting single image depth
  estimation: Toward higher resolution maps with accurate object boundaries,''
  in \emph{2019 IEEE Winter Conference on Applications of Computer Vision
  (WACV)}.\hskip 1em plus 0.5em minus 0.4em\relax IEEE, 2019, pp. 1043--1051.

\bibitem{alhashim2018high}
I.~Alhashim and P.~Wonka, ``High quality monocular depth estimation via
  transfer learning,'' \emph{arXiv preprint arXiv:1812.11941}, 2018.

\bibitem{Ummenhofer_2017}
B.~Ummenhofer, H.~Zhou, J.~Uhrig, N.~Mayer, E.~Ilg, A.~Dosovitskiy, and
  T.~Brox, ``Demon: Depth and motion network for learning monocular stereo,''
  \emph{2017 IEEE Conference on Computer Vision and Pattern Recognition
  (CVPR)}, Jul 2017.

\bibitem{wang2004image}
Z.~Wang, A.~C. Bovik, H.~R. Sheikh, and E.~P. Simoncelli, ``Image quality
  assessment: from error visibility to structural similarity,'' \emph{IEEE
  transactions on image processing}, vol.~13, no.~4, pp. 600--612, 2004.

\bibitem{2014Adam}
D.~P. Kingma and J.~Ba, ``Adam: A method for stochastic optimization,''
  \emph{CoRR}, vol. abs/1412.6980, 2015.

\bibitem{9130033}
Y.~{Yao}, M.~{Roxas}, R.~{Ishikawa}, S.~{Ando}, J.~{Shimamura}, and T.~{Oishi},
  ``Discontinuous and smooth depth completion with binary anisotropic diffusion
  tensor,'' \emph{IEEE Robotics and Automation Letters}, vol.~5, no.~4, pp.
  5128--5135, 2020.

\bibitem{schuster2021ssgp}
R.~Schuster, O.~Wasenmuller, C.~Unger, and D.~Stricker, ``Ssgp: Sparse spatial
  guided propagation for robust and generic interpolation,'' in
  \emph{Proceedings of the IEEE/CVF Winter Conference on Applications of
  Computer Vision (WACV)}, January 2021, pp. 197--206.

\bibitem{Eldesokey_2020_CVPR}
A.~Eldesokey, M.~Felsberg, K.~Holmquist, and M.~Persson, ``Uncertainty-aware
  cnns for depth completion: Uncertainty from beginning to end,'' in
  \emph{IEEE/CVF Conference on Computer Vision and Pattern Recognition (CVPR)},
  June 2020.

\bibitem{Menze2018JPRS}
M.~Menze, C.~Heipke, and A.~Geiger, ``Object scene flow,'' \emph{ISPRS Journal
  of Photogrammetry and Remote Sensing}, vol. 140, pp. 60--76, 2018, geospatial
  Computer Vision.

\bibitem{Silberman:ECCV12}
N.~Silberman, D.~Hoiem, P.~Kohli, and R.~Fergus, ``Indoor segmentation and
  support inference from rgbd images,'' in \emph{Computer Vision -- ECCV 2012},
  A.~Fitzgibbon, S.~Lazebnik, P.~Perona, Y.~Sato, and C.~Schmid, Eds.\hskip 1em
  plus 0.5em minus 0.4em\relax Berlin, Heidelberg: Springer Berlin Heidelberg,
  2012, pp. 746--760.

\bibitem{qi2018frustum}
C.~R. Qi, W.~Liu, C.~Wu, H.~Su, and L.~J. Guibas, ``Frustum pointnets for 3d
  object detection from rgb-d data,'' in \emph{Proceedings of the IEEE
  Conference on Computer Vision and Pattern Recognition}, 2018, pp. 918--927.

\bibitem{chen20153d}
X.~Chen, K.~Kundu, Y.~Zhu, A.~G. Berneshawi, H.~Ma, S.~Fidler, and R.~Urtasun,
  ``3d object proposals for accurate object class detection,'' in
  \emph{Advances in Neural Information Processing Systems}, 2015, pp. 424--432.

\bibitem{Geiger2012CVPR}
A.~{Geiger}, P.~{Lenz}, and R.~{Urtasun}, ``Are we ready for autonomous
  driving? the kitti vision benchmark suite,'' in \emph{2012 IEEE Conference on
  Computer Vision and Pattern Recognition}, 2012, pp. 3354--3361.

\bibitem{murORB2}
R.~Mur-Artal and J.~D. Tard\'os, ``{ORB-SLAM2}: an open-source {SLAM} system
  for monocular, stereo and {RGB-D} cameras,'' \emph{IEEE Transactions on
  Robotics}, vol.~33, no.~5, pp. 1255--1262, 2017.

\bibitem{murTRO2015}
Mur-Artal, Ra\'ul, Montiel, J.~M. M., and J.~D. Tard\'os, ``{ORB-SLAM}: a
  versatile and accurate monocular {SLAM} system,'' \emph{IEEE Transactions on
  Robotics}, vol.~31, no.~5, pp. 1147--1163, 2015.

\end{thebibliography}

\end{document}